\documentclass[10pt,twocolumn,letterpaper]{article}

\usepackage[pagenumbers]{cvpr} 

\definecolor{cvprblue}{rgb}{0.21,0.49,0.74}
\usepackage[pagebackref,breaklinks,colorlinks,allcolors=cvprblue]{hyperref}
\usepackage{bm}
\usepackage{amsmath}
\usepackage{amssymb}
\usepackage{multirow}
\usepackage{colortbl}
\usepackage{booktabs}
\usepackage{xcolor}
\usepackage{soul}
\sethlcolor{yellow} 
\usepackage{pifont}

\def\paperID{17093} 
\def\confName{CVPR}
\def\confYear{2026}

\title{IrisNet: Infrared Image Status Awareness Meta Decoder for \\ Infrared Small Targets Detection}

\author{Xuelin Qian \and
Jiaming Lu \and
Zixuan Wang \and
Wenxuan Wang \and
Zhongling Huang \and
Dingwen Zhang \and
Junwei Han \and
\textit{Northwestern Polytechnical University} \\
{\tt\small \{xlqian, jmlu, wzixuan, wxwang, huangzhongling, zdw2006yyy, jhan\}@nwpu.edu.cn}
}

\begin{document}
\maketitle

\begin{abstract}
Infrared Small Target Detection (IRSTD) faces significant challenges due to low signal-to-noise ratios, complex backgrounds, and the absence of discernible target features. While deep learning-based encoder-decoder frameworks have advanced the field, their static pattern learning suffers from pattern drift across diverse scenarios (\emph{e.g.}, day/night variations, sky/maritime/ground domains), limiting robustness. To address this, we propose IrisNet, a novel meta-learned framework that dynamically adapts detection strategies to the input infrared image status. Our approach establishes a dynamic mapping between infrared image features and entire decoder parameters via an image-to-decoder transformer. More concretely, we represent the parameterized decoder as a structured 2D tensor preserving hierarchical layer correlations and enable the transformer to model inter-layer dependencies through self-attention while generating adaptive decoding patterns via cross-attention. To further enhance the perception ability of infrared images, we integrate high-frequency components to supplement target-position and scene-edge information. Experiments on NUDT-SIRST, NUAA-SIRST, and IRSTD-1K datasets demonstrate the superiority of our IrisNet, achieving state-of-the-art performance.
\end{abstract}

\section{Introduction}
\label{sec:intro}

Infrared Small Target Detection (IRSTD) focuses on detecting and localizing small objects in infrared images captured by satellites, aircraft, or ground‑based sensors~\cite{IRSTDsurvey}. These targets are typically tiny, textureless regions buried in cluttered, dynamic backgrounds with severe noise interference, resulting in challenges such as low signal‑to‑noise ratios, weak contrast, and indistinct structural features~\cite{NetSurvey}. Early IRSTD research relied on handcrafted priors, including low‑rank representations~\cite{IPI}, human visual system models~\cite{ALCM}, and filtering-based enhancements~\cite{Top-hat}. While computationally efficient, these approaches struggle with robustness in complex real-world scenes. Nevertheless, IRSTD remains critical for applications in military reconnaissance, public security, and industrial inspection~\cite{sea_assist,publicsecurity}, driving continued research interest.

\begin{figure}[t]
    \centering
    \includegraphics[width=1.0\linewidth]{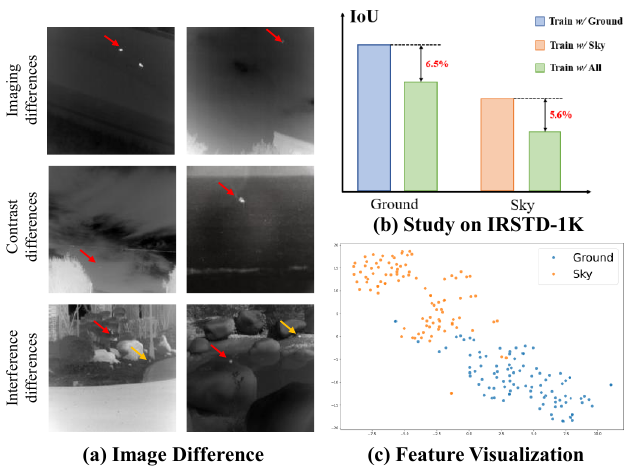} 
    \vspace{-0.1in}
    \caption{\textbf{Pattern drift analysis.} (a) Infrared targets can exhibit different characteristics across different scenarios. The red arrows represent the target, and the yellow arrows represent false alarms. (b) Our pilot study demonstrates that adopting a static decoding paradigm for different scenarios may result in suboptimal IRSTD performance due to the issue of pattern shift. (c) Visualizations of feature distribution to show the discrepancy between infrared images with different targets and scenarios. Thereby, it requires distinct decoding patterns to localize them.  
    }     
    \label{fig:intro}
    \vspace{-0.15in}
\end{figure}

With the rise of deep learning, encoder–decoder frameworks have become dominant in modern IRSTD methods~\cite{UNet}. Encoders extract low-level textures and high-level semantics, while decoders reconstruct pixel-level masks. To enhance small target perception, recent studies incorporate multi‑scale receptive fields, salient-region emphasis, and contextual feature fusion. For instance, SCTrans~\cite{Sctransnet} employs spatial–channel transformers to distinguish targets from similar backgrounds; MDAFNet~\cite{MDAFNet} promotes feature reuse via dense inter-layer connections; and DCFR‑Net~\cite{DCFRNet} leverages pixel intensity and directional differences to enhance target detail.

Despite recent progress, most methods focus only on pattern learning and overlook the performance degradation caused by pattern drift across diverse scenarios. As illustrated in Fig.~\ref{fig:intro}(a), infrared targets exhibit widely varying appearances: a target may appear bright in one situation and dim in another; sky scenes can be disrupted by clouds, while maritime scenes offer clearer contrast; and ground backgrounds often contain distractors with similar visual properties (\emph{e.g.}, small stones in grass). These heterogeneous imaging characteristics lead to inconsistent target features, forming the essence of pattern drift. To quantify its impact, we conduct two analyses.  First, cross-scenario evaluations (Fig.~\ref{fig:intro}(b)) show that static models (\emph{e.g.}, U-Net) trained on mixed data degrade notably when tested on a specific scenario—IoU drops by 8–12\% and false alarms rise by 10–30\%—indicating their inability to accommodate varying feature patterns. Second, t-SNE visualization (Fig.~\ref{fig:intro}(c)) further highlights this issue, as features from different scenarios form clearly separated clusters, reflecting substantial shifts in target representations. These results confirm that static models learn a compromised pattern that fails to adapt to scenario-specific variations, leading to suboptimal localization and increased false detections.

To address the challenge of pattern drift in IRSTD, we propose the InfRared Image Status awareness Network (IrisNet), which dynamically maps infrared image status—such as scene context and platform type—to decoder configurations. Our core innovation lies in a meta-learned decoding paradigm, where features extracted by the encoder dynamically generate the decoder parameters. This adaptive mechanism enables the model to tailor its detection strategy according to the unique discriminative patterns present in each input image, effectively mitigating the limitations of static learned patterns.

Directly predicting the entire decoder parameter vector risks disrupting the hierarchical relationships among parameters. To overcome this, we represent the decoder parameters as a structured 2D tensor, with rows corresponding to decoder layers and columns to parameters within each layer. Treating this tensor as a learnable query, we design an image-to-decoder querying transformer that preserves structural information while explicitly modeling inter-layer parameter correlations. Unlike conventional dynamic convolution approaches that adapt only a few layers, our meta-decoder covers all decoder layers and supports advanced components such as multi-scale and spatial attention blocks. Additionally, we integrate high-frequency components into the encoder to enrich edge information and potential target locations, enhancing the encoder’s awareness of the infrared image and improving the meta-decoder’s adaptability.

In summary, IrisNet employs an encoder to extract image perception features and an image-to-decoder transformer that generates the full set of decoder parameters conditioned on these features. Cross-attention enables feature conditioning, while self-attention captures parameter relationships across layers. The generated tokens are rearranged to form a fully parameterized decoder that dynamically interprets target masks based on input image status. \textbf{Our contributions can be summarized as follows:}
\textbf{(1)} We introduce IrisNet, a novel framework with the meta-learned decoding paradigm to address the pattern drift issue in IRSTD. To our knowledge, we are the first to establish a dynamic mapping between the infrared image status and the entire decoder.
\textbf{(2)} We propose an image-to-decoder transformer with a structured decoder representation, which enables the correlation dependencies between parameters of hierarchical layers, facilitating the diversity, stability, and adaptability of the meta decoder.
\textbf{(3)} Extensive experiments on NUDT-SIRST, NUAA-SIRST, and IRSTD-1K datasets demonstrate that IrisNet outperforms existing methods, achieving new state-of-the-art performance.

\section{Related Work}
\label{sec:related_work}

\subsection{Infrared Small Target Detection}
In IRSTD, research has progressed from conventional image processing approaches—such as background modeling, local contrast enhancement, and filter-based techniques like low-rank decomposition for background suppression~\cite{PSTNN} and spatial filters leveraging intensity gradients~\cite{TOPHAT2}—to deep learning frameworks that dominate current advancements~\cite{NetSurvey}. Convolutional neural networks (CNNs) have shown strong performance by aligning with the local saliency of small targets, as seen in models like ACMNet~\cite{ACMNet} (asymmetric context modulation), DNANet~\cite{DNANet} (hierarchical feature interaction), ALCNet~\cite{ALCNet} (local contrast enhancement), UIU-Net~\cite{UIUNet} (nested U-Net for global-local fusion), ISNet~\cite{ISNet} (edge modeling via Taylor finite difference), and Dim2Clear~\cite{Dim2Clear} (super-resolution and enhancement). However, these CNNs are constrained by local receptive fields, limiting their ability to capture long-range contextual dependencies critical for distinguishing targets from cluttered or low-contrast backgrounds. This limitation has driven growing interest in transformer-based architectures, which provide stronger global context modeling and improved robustness for complex IRSTD scenarios.

\begin{figure*}[t]
    \centering
    \includegraphics[width=1.\textwidth]{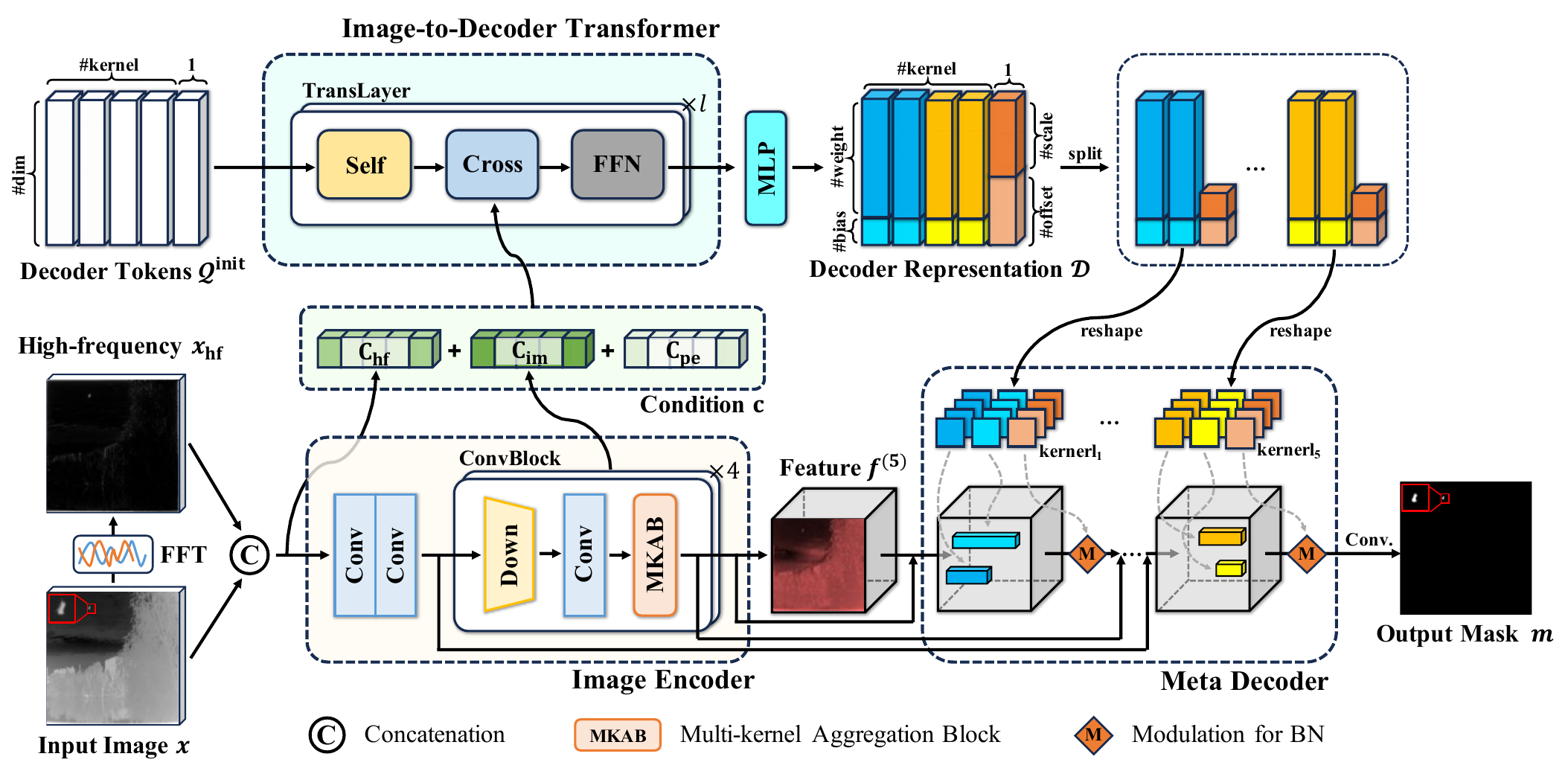} 
    \caption{Overview of the proposed IrisNet architecture. (1) Image Encoder extracts high-frequency–enhanced hierarchical features; (2) Image-to-Decoder transformer maps features to decoder parameters through learnable tokens; (3) Meta Decoder constructs the decoder to output binary localization masks. }
    \label{fig:framework}  
    \vspace{-0.1in}
\end{figure*}

\subsection{Transformer methods in IRSTD}
The Vision Transformer (ViT)~\cite{ViT} models images as patch sequences via self-attention, enabling robust long-range dependency modeling, and has been widely adopted in IRSTD for global context capture~\cite{TransSurvey}. Building on this, RKFormer~\cite{Rkformer} combines Runge–Kutta blocks with random-connection attention to balance global and local features, IAANet~\cite{IAANet} employs a coarse-to-fine encoder for region-level pixel relations, TCI-Former~\cite{TCIFormer} integrates thermal-conduction-inspired attention with boundary refinement, and SCTransNet~\cite{Sctransnet} fuses multi-level semantics with spatial cues through spatial–channel cross-transformer blocks. While these methods enhance global modeling and boundary refinement, they remain encoder-centric and lack input-adaptive decoding. Our IrisNet addresses this by introducing a meta-learned decoder that dynamically generates parameters conditioned on image status, enabling adaptive detection across diverse scenarios.

\subsection{Hyper Network}

Hypernetworks, first introduced by Ha et al.~\cite{2016Hypernetworks}, break from the traditional fixed-parameter paradigm by allowing a meta-network to dynamically generate the weights of a target network. Initially proposed for image classification, they have since been extended to diverse vision tasks. For example, HyperSeg~\cite{Hyperseg} leverages a hypernetwork-based encoder to generate decoder parameters for semantic segmentation, while CVAE-H~\cite{Cvae-H} conditions network heads on instance-level priors for autonomous driving. Transformer-based hypernetworks advance this concept by enabling set-to-set mappings, enhancing expressiveness for multi-modal conditioning~\cite{ECCV22Hyper}. Further, HyperNeRF~\cite{HyperNeRF} employs diffusion-driven hypernetworks to regress neural radiance field weights without per-scene optimization, and HyperFields~\cite{HyperFields} generalizes this approach to text-conditioned zero-shot NeRF generation using Transformer-based cross-modal weight generators.
In infrared small target detection (IRSTD), hypernetworks remain underexplored, especially for decoders integrating multi-scale features. We introduce a Transformer-conditioned hypernetwork that predicts all decoder layer parameters in a single pass, representing them as a structured 2D tensor to preserve inter-layer correlations. This design makes the decoder sample-adaptive and structure-aware, effectively mitigating pattern drift across diverse scenarios.

\section{Methodology}
\label{sec:method}

In this section, we elaborate on the proposed IrisNet, which takes infrared images as inputs and outputs the binary masks to localize small targets. Figure~\ref{fig:framework} illustrates the schematic overview of our framework, consisting of three components. 
The \textit{image encoder} first encodes the input image into hierarchical features, enhanced by high-frequency information.
Then, the \textit{image-to-decoder transformer} establishes a dynamic mapping from image features to the decoder configuration via cross-attention. This configuration is represented as learnable tokens, and then projected into the parameter space.
Lastly, the \textit{meta decoder} reorganizes the output tokens to build a parameterized decoder with functional layers.

\subsection{Image Encoder \label{sec:image_encoder}}

Given an infrared image $\bm{x}$ as input, IrisNet first applies the encoder to extract multi-scale features $\left\{ f^{(i)}  \right\}_{i=1}^{n} \in \mathbb{R}^{H_{i} \times W_{i} \times C_{i}}$ from the shallow to the deep layer, where $H_{i}$ and $W_{i}$ denote the height and width of the $i$-th feature map and $C_{i}$ is the corresponding feature dimension. Specifically, our encoder adopts five convolution blocks ($n=5$), each of which has two stacked convolution layers to process features. For efficiency, we gradually reduce the size of the feature map by half from the second block. Intuitively, a stronger encoder can capture more image status information, which is beneficial for producing a decoding paradigm that is more suitable for the input. 
Consequently, we additionally incorporate two designs of frequency domain and large receptive fields into the encoder.

\noindent \textbf{Frequency-assisted Encoder.} Unlike visible images, where edges are mainly defined by texture, edges in infrared images are more strongly influenced by temperature variations. High-pass filtering effectively captures this physical distinction. By introducing high-frequency signals in the frequency domain, we can enhance the visibility of targets while simultaneously suppressing background thermal noise. This is expected to improve the perception of scene contours.
Concretely, we first define a Gaussian mask $\mathcal{M}_{hp}$ as high-pass filtering, which is expressed as,
\begin{equation}
\mathcal{M}_{\text{hp}}(u, v) = 1 - \exp\left(-\frac{(u-u_0)^2 + (v-v_0)^2}{2\sigma^2}\right)
\end{equation}
\noindent where $\left(u, v\right)$ represents frequency coordinates, $\left(u_0, v_0\right)$ is the center of the frequency spectrum, and $\sigma$ controls the filter's cutoff frequency. Subsequently, we perform frequency domain processing on the input image and apply the Gaussian filtering mask,
\begin{equation}
\bm{x}_{\text{hp}}=\mathcal{F}^{-1}\left(\mathcal{M}_{\text{hp}} \odot \mathcal{F}\left(\bm{x}\right)\right)
\end{equation}
\noindent where $\mathcal{F}$ and $\mathcal{F}^{-1}$ denote the forward and inverse Fourier transforms, respectively. After obtaining the high-frequency signal $\bm{x}_{\text{hp}}$, we concatenate it with the input image along the channel dimension as $\bm{x}_{\text{sf}}$, and pass them through the encoder to capture multi-level feature representations.

\begin{figure}[t]
    \centering
    \includegraphics[width=1\linewidth]{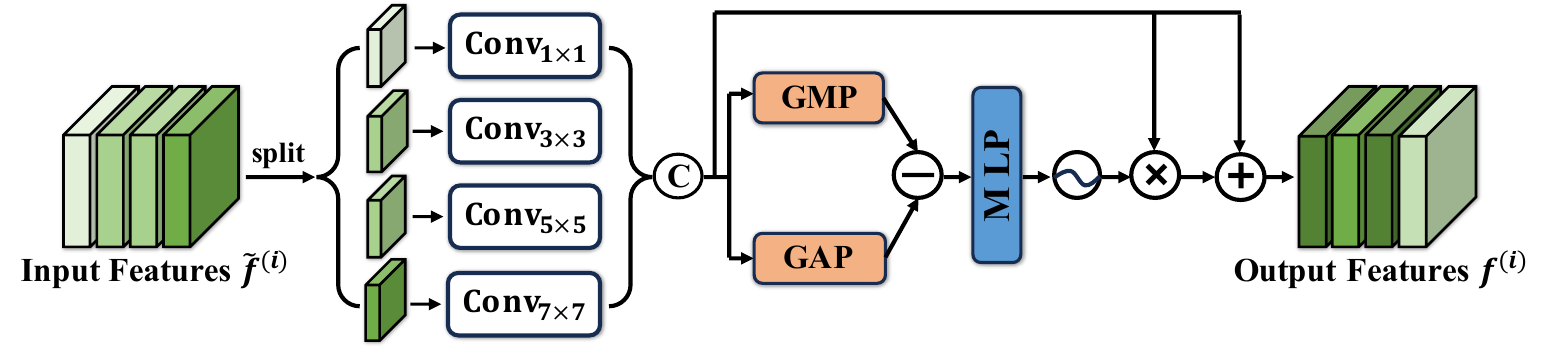}
    \caption{Details of Multi-kernel Aggregation Block (MKAB).}
    \label{fig:MKAB}     
    \vspace{-0.1in}
\end{figure}

\noindent \textbf{Multi-kernel Aggregation Block.} 
Small targets usually occupy only a few pixels in the image, which could especially be suppressed during feature downsampling. To better perceive them in a noisy background, we introduce the Multi-kernel Aggregation Block (MKAB), placed at the end of each convolution block, capturing contextual information through parallel paths with varying receptive fields. 
As shown in Fig.~\ref{fig:MKAB}, it splits the input channel into parallel groups, each of which is then processed by a convolution with a different kernel size. Nevertheless, the output results from these parallel paths may contain redundancies, therefore, we further append a channel attention (CA) to emphasize semantically significant channels while suppressing less relevant ones. The overall process can be formulated as,
\begin{equation}
\begin{aligned}
&\tilde{f}^{\left(i\right)} = \oplus_{k \in \mathcal{K} } \text{Conv}_{(2k-1)\times (2k-1)}(\tilde{f}_{k}^{\left(i\right)})  \\
f^{\left(i\right)} = \tilde{f}^{\left(i\right)}& \cdot (1 + \sigma(\text{MLP}(\text{GMP}(\tilde{f}^{\left(i\right)})-\text{GAP}(\tilde{f}^{\left(i\right)}))) )
\end{aligned}
\end{equation}

\noindent where $\mathcal{K}=\left \{ 1,2,3,4 \right \} $ is the set of kernel sizes, $\tilde{f}_{k}^{\left(i\right)}$ represents the $k$-th channel group of the intermediate features $\tilde{f}^{\left(i\right)}$ in the $i$-th block, $\oplus$ means the channel-wise concatenation, GMP and GAP denote global max and average pooling, and MLP is the multilayer perceptron. 

\subsection{Image-to-decoder Transformer \label{sec:transformer}}

Infrared imaging is fundamentally based on thermal radiation, which means that the same target can appear different or even opposite characteristics depending on factors such as background or time of capture (as illustrated in Fig.~\ref{fig:intro}). Instead of training one decoder for all images, we implement an image-to-decoder transformer that dynamically builds a tailored decoder for each input. This transformer can be regarded as a meta-learned decoding paradigm that learns to translate image features to decoder parameters in a data-driven way.

\noindent \textbf{Image-awareness Condition.} 
Conditioning on image features is a necessary part of fulfilling the image status awareness decoding that we advocate. We construct the condition $\mathbf{c}$ of the input images by integrating their multi-scale features. Similar to ViT~\cite{ViT}, a convolution layer with specified kernel size and stride is used to project multi-scale features into patch-wise tokens $ \mathbf{c}_\text{im} = \left[ c_{1}(f^{(2)}); c_{2}(f^{(3)}); c_{3}(f^{(4)}); c_{4}(f^{(5)}) \right] \in \mathbb{R}^{L\times C_{T}}$, where $\left\{c_{i}\right\}_{i=1}^{4}$ denotes convolution layers, $C_{T}$ is the hidden dimension, and $L$ means the token length. 
To further enhance spatial perception, we project $\bm{x}_\text{sf}$ into high-frequency tokens $\mathbf{c}_\text{hf}$. In addition, we incorporate the cosine-sine positional embedding $\mathbf{c}_\text{pe}$ to preserve the spatial information. In the end, we combine all components by summing them element-wise to obtain the final image-aware condition $ \mathbf{c} = \mathbf{c}_\text{im} + \mathbf{c}_\text{hf} + \mathbf{c}_\text{pe}$. This captures both hierarchical semantic features and explicit high-frequency details, resulting in a comprehensive representation of the image status.

\noindent \textbf{Decoder Representation.}
The parameters of a convolutional layer\footnote{For simplicity, we omit \textit{group} and \textit{bias} here.} are typically structured as a 4D tensor with shape $\left(C_{\text{out}}, C_{\text{in}}, k, k \right)$, where $C_{\text{out}}$ and $C_{\text{in}}$ denote the number of output kernels and input channels, respectively, and $k$ is the kernel size.
Prior works often flatten these parameters into a single vector for prediction~\cite{2016Hypernetworks}. However, this approach significantly increases the cost (\textit{e.g.}, expanding from $C_{\text{out}}$ to $C_{\text{out}} \times C_{\text{in}} \times k^2$) and disrupts the semantic structure of the parameters across dimensions.

In contrast, we maintain a structured 2D tensor representation of the decoder, as illustrated at the top of Fig.~\ref{fig:framework}. Specifically, we standardize the number of input channels across all decoder layers and apply a factorization principle (\textit{i.e.}, multiplying like terms) to preserve the original parameter hierarchy while reducing computational complexity. Assuming that the decoder consists of three layers, then our decoder representation is,
\begin{equation}
\begin{aligned}
C_{\text{out,1}} \times  C_{\text{in}}& \times k^2 + C_{\text{out,2}} \times C_{\text{in}} \times k^2 + C_{\text{out,3}} \times C_{\text{in}} \times k^2 \\
&= \left(C_{\text{out,1}} + C_{\text{out,2}} + C_{\text{out,3}} \right) \times \left( C_{\text{in}} \times k^2 \right)
\end{aligned}
\end{equation}
\noindent where the subscripts $[1, 2, 3]$ indicate the corresponding decoder layers. The first item stands for the total number of convolution kernels for all layers, while the second item corresponds to the number of parameters per kernel. In our implementation, we set $C_{out} = C_{in}$. 
Since batch normalization (BN) plays a key role in convolutional networks by adjusting feature distributions~\cite{BN}, we also incorporate it into our meta-learned decoder. To do so, we allocate an additional column in the parameter tensor to represent the BN parameters for all decoder layers.

Such a decoder representation endows two advantages: \textbf{(1)} Its row-wise organization encodes the vertical structural correlation, maintaining the sequential dependency between decoder layers. \textbf{(2)} Its column-wise organization enforces horizontal parameter consistency. This enables the model to learn systematic patterns in its variation with layer depth.

\noindent \textbf{Transformer Layers.}
To obtain the decoder representation $\mathcal{D}$, we initially define a set of learnable decoder tokens $\mathbf{Q^{\text{init}}} \in \mathbb{R}^{(N_{q}+1) \times C_{T}}$, where $N_{q}$ is the number of tokens (\textit{i.e.}, the total number of kernels) and the extra ``1'' is the norm token responsible for BN representation. We adopt the Querying Transformer ~\cite{BLIP} to iteratively refine these tokens by incorporating information from the image condition $\mathbf{c}$
After passing through all transformer layers, we obtain the refined decoder tokens $\mathbf{Q^{\text{out}}}$ from the last layer. A projection layer is further applied to adjust their dimension, producing the final decoder representation $\mathcal{D}$.

Each transformer layer consists of a self-attention layer, a cross-attention layer and a feed-forward network.
Firstly, the self-attention layer captures the relationships among the decoder tokens, modeling dependencies across the hierarchical structure of the decoder.
Next, the updated tokens are passed through the cross-attention layer, which builds alignments and interactions between the image features and the decoder tokens, allowing the model to condition the decoding pattern on the input image. 
Finally, a standard feed-forward network is applied to perform nonlinear transformations, enhancing the capacity of token feature learning. 
The overall operations within each transformer layer can be summarized as follows,
\begin{equation}
\begin{aligned}
f_\text{self}^\text{in} &= \text{SelfAttn}( f^\text{in},f^\text{in},f^\text{in} ) + f^\text{in} \\
f_\text{cross}^\text{in} &= \text{CrossAttn}( f_\text{self}^\text{in}, \mathbf{c},\mathbf{c} ) + f_\text{self}^\text{in} \\
f^\text{out} &= \text{FFN}(f_\text{cross}^\text{in}) + f_\text{cross}^\text{in}
\end{aligned}
\end{equation}
\noindent where $f^\text{in}$ and $f^\text{out}$ represent the input and output features of the transformer layer, respectively.

\begin{figure}[t]
    \centering
    \includegraphics[width=\linewidth]{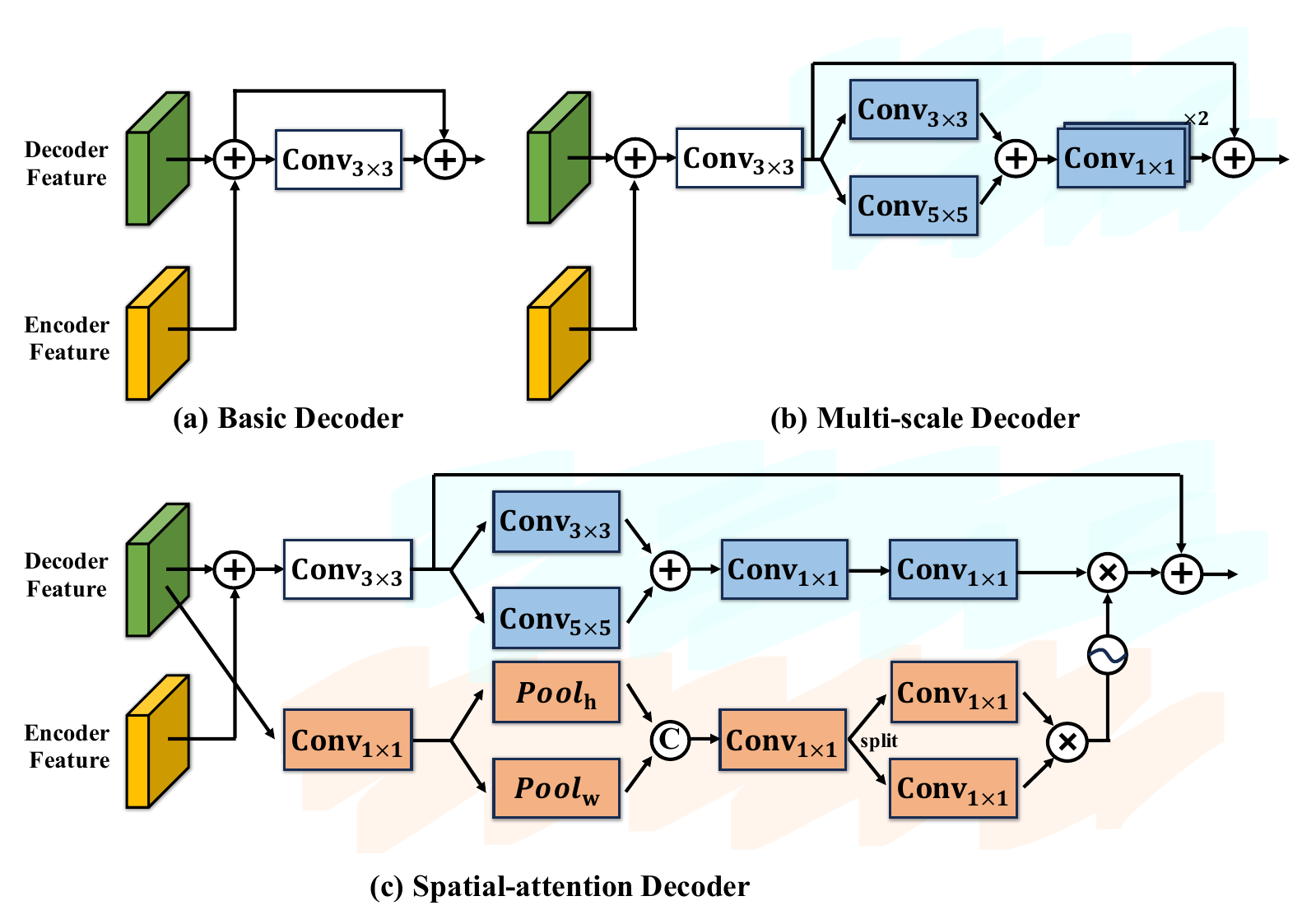} 
    \caption{Details of three types of Meta Decoder. All parameters are dynamically produced by the Image-to-decoder transformer. (a) uses stacked $3\times3$ conv + BatchNorm + ReLU layers for efficient feature extraction. (b) Additionally, it adds parallel $3\times3$ and $5\times5$ depthwise separable convolutions to capture multi-scale features. (c) Further augments the multi-scale design with a spatial attention branch to reinforce targets and suppress background.}         
    \label{fig:metadecoder}   
    \vspace{-0.1in}
\end{figure}

\begin{table*}[t]
  \centering
  \footnotesize
    \setlength{\tabcolsep}{3.2mm}{
    \begin{tabular}{lccccccccccc}
    \toprule
    \multicolumn{1}{c}{\multirow{2}{*}{\textsc{Method}}} & \multirow{2}{*}{\textsc{Backbone}} & \multirow{2}{*}{\textsc{Size}} & \multicolumn{3}{c}{\textbf{IRSTD-1K}} & \multicolumn{3}{c}{\textbf{NUDT-SIRST}} & \multicolumn{3}{c}{\textbf{NUAA-SIRST}} \\
    \cmidrule(r){4-6} \cmidrule(r){7-9} \cmidrule(r){10-12}
    & & & IoU $\uparrow$ & P$_{d} \uparrow$ & F$_{a} \downarrow$ & IoU $\uparrow$ & P$_{d} \uparrow$ & F$_{a} \downarrow$ & IoU $\uparrow$ & P$_{d} \uparrow$ & F$_{a} \downarrow$ \\
    \midrule
    Top-Hat~\cite{Top-hat} & Traditional & 256 & 10.06 & 75.11 & 1432 & 20.72 & 78.41 & 166.7 & 7.143 & 79.84 & 1012 \\
    PSTNN~\cite{PSTNN} & Traditional & 256 & 24.57 & 71.99 & 35.26 & 14.85 & 66.13 & 44.17 & 30.30 & 72.80 & 48.99 \\
    MSLSTIPT~\cite{MSLSTIPT} & Traditional & 256 & 11.43 & 79.03 & 1524 & 8.342 & 47.40 & 888.1 & 10.30 & 82.13 & 1131 \\
    \midrule
    \rowcolor{gray!10} RKformer~\cite{Rkformer} & Hybrid & 512 & 64.12 & 93.27 & 18.65 & 92.25 & 96.86 & 6.580 & 77.24 & 99.11 & \textbf{1.580} \\
    \rowcolor{gray!10} TCI-Former$^{\dag}$~\cite{TCIFormer} & Hybrid & 512 & \underline{70.14} & \textbf{96.31} & 14.81 & - & - & - & \underline{80.79} & 99.23 & 4.189  \\
    IAANet~\cite{IAANet} & Hybrid & 256 & 66.25 & 93.15 & 14.20 & 90.22 & 97.26 & 8.320 & 74.22 & 93.53 & 22.70 \\
    MTU-Net~\cite{MTUNet} & Hybrid & 256 & 66.11 & 93.27 & 36.80 & 74.85 & 93.97 & 46.95 & 74.78 & 93.54 & 22.36 \\
    SCTransNet~\cite{Sctransnet} & Hybrid & 256 & 68.03 & 93.27 & \underline{10.74} & \underline{94.09} & 98.62 & \underline{4.290} & 77.50 & 96.95 & 13.92 \\
    \midrule
    \rowcolor{gray!10} IRPruneDet$^{\dag}$~\cite{IRPruneDet} & CNN & 512 & 64.54 & 91.74 & 16.04 & - & - & - & 75.12 & 98.61 & 2.960 \\
    \rowcolor{gray!10} GCI-Net~\cite{GCI-Net} & CNN & 512 & 67.75 & 93.89 & 12.84 & 92.43 & 98.25 & 8.960 & 78.81 & \underline{99.34} & \underline{2.110} \\
    \rowcolor{gray!10} CFD-Net$^{\dag}$~\cite{CFD-Net} & CNN & 512 & 69.97 & \underline{95.96} & 14.21 & - & - & - & 80.56 & \textbf{99.58} & 2.880 \\
    ACM~\cite{ACMNet} & CNN & 256 & 59.23 & 93.27 & 65.28 & 61.12 & 93.12 & 55.22 & 68.93 & 91.63 & 15.23 \\
    ALCNet~\cite{ALCNet} & CNN & 256 & 60.60 & 92.98 & 58.80 & 64.74 & 94.18 & 34.61 & 70.83 & 94.30 & 36.15 \\
    ISNet~\cite{ISNet} & CNN & 256 & 61.85 & 90.24 & 31.56 & 81.24 & 97.78 & 6.340 & 70.49 & 95.06 & 67.98 \\
    FC3-Net~\cite{FC3-Net} & CNN & 256 & 65.07 & 91.54 & 15.55 & 78.56 & 93.86 & 23.92 & 72.44 & 98.14 & 10.85 \\
    DNA-Net~\cite{DNANet} & CNN & 256 & 65.90 & 90.91 & 12.24 & 88.19 & \underline{98.83} & 9.000 & 75.80 & 95.82 & 8.780 \\
    UIU-Net~\cite{UIUNet} & CNN & 256 & 66.15 & 93.98 & 22.07 & 93.48 & 98.31 & 7.790 & 76.91 & 95.82 & 14.13 \\
    ISTDU~\cite{ISTDUNet} & CNN & 256 & 66.36 & 93.60 & 53.10 & 89.55 & 97.67 & 13.44 & 75.52 & 96.58 & 14.54 \\
    RDIAN~\cite{RDIAN} & CNN & 256 & 56.45 & 88.55 & 26.63 & 76.28 & 95.77 & 34.56 & 68.72 & 93.54 & 43.29 \\
    AGPCNet~\cite{AGPCNet} & CNN & 256 & 66.29 & 92.83 & 13.12 & 88.87 & 97.20 & 10.02 & 75.69 & 96.48 & 14.99 \\
    Dim2Clear~\cite{Dim2Clear} & CNN & 256 & 66.34 & 93.75 & 20.93 & 81.37 & 96.23 & 9.170 & 77.29 & 99.10 & 6.720 \\
    MMLNet~\cite{MMLNet} & CNN & 256 & 67.21 & 94.28 & 14.00 & 81.81 & 98.43 & 11.77 & 78.71 & 98.88 & 25.71 \\
    L2SKNet~\cite{L2SKNet} & CNN & 256 & 67.81 & 90.24 & 17.46 & 93.58 & 97.57 & 5.330 & 73.43 & 98.17 &  20.82 \\
    \midrule
    IrisNet (\textit{Ours}) & CNN & 256 & \textbf{72.08} & 92.59 & \textbf{9.641} & \textbf{95.74} & \textbf{99.37} & \textbf{0.367} & \textbf{81.77} & 97.34 & 9.392 \\
    \bottomrule
    \end{tabular}}
    \caption{Comparison with existing IRSTD approaches on the IRSTD-1K, NUDT-SIRST, and NUAA-SIRST datasets. The evaluation metrics are IoU ($10^{-2}$), $P_d$ ($10^{-2}$), and $F_a$ ($10^{-6}$). Note that methods in the gray area differ from others in the experimental setting. They either use $512 \times 512$ images, or apply a different training/testing split (marked with ``$\dag$''), or both. 
    The first and second best results are highlighted in \textbf{bold} and \underline{underlined}, respectively.}
  \label{tab:sota_1}
\end{table*}

\subsection{Meta Decoder\label{sec:meta-decoder}}

By establishing a dynamic mapping between image status and decoder configuration, IrisNet can adaptively choose the most suitable decoding pattern for each input. Once the decoder representation $\mathcal{D}$ is obtained, it is reshaped into a parameterized meta-decoder $\mathcal{D}^\text{meta}$, which takes the image features $f^{(5)}$ as input and processes them layer by layer, as illustrated in the right part of Fig.~\ref{fig:framework}. To preserve fine-grained details, we follow previous works to incorporate skip connections between the encoder and the meta-decoder, reusing the low-level features $\left\{ f^{(i)} \right\}_{i=1}^{4}$. Finally, a shared convolution layer is applied to produce the final target mask as a single-channel output.

Crucially, thanks to the compact and structured decoder representation, we can flexibly learn different mapping strategies to construct meta-decoders with varying functions and capacities. 
In our implementation, we construct three types of meta-decoders, ranging from simple to advanced configurations, as illustrated in Fig.~\ref{fig:metadecoder}.
\begin{itemize}
    \item \textbf{Basic Decoder:} Each decoder layer consists of a single $3 \times 3$ convolution layer followed by Batch Normalization and a ReLU activation. It offers foundational feature extraction while maintaining computational efficiency. 
    \item \textbf{Multi-scale Decoder:} Building on the basic decoder, we further introduce a multi-scale module to increase the depth of each decoding layer. This module applies both $3 \times 3$ and $5 \times 5$ depth-wise convolutions to capture multi-scale spatial features. The outputs are summed and then passed through two point-wise convolution layers for information compression and restoration. This design allows the model to efficiently capture features across a range of receptive fields.
    \item \textbf{Spatial-attention Decoder:} Building on the multi-scale decoder, we also introduce a position attention branch~\cite{ASFYOLO} that captures spatial attention by separately modeling information along the vertical and horizontal directions. This design helps enhance target features while suppressing background noise in the outputs of the multi-scale branch.
\end{itemize}

\subsection{Training Objective \label{sec:loss}}

IrisNet produces the predicted target mask $\bm{m}$ from an input infrared image and leverages the corresponding ground-truth mask $\bm{\hat{m}}$ as supervision during training. More precisely, for every input image $\bm{x}$, the training objective is to minimize the following loss function, 
\begin{equation}
\mathcal{L}_{\text{total}} = \mathcal{L}_{\text{BCE}}\left(\bm{m}, \bm{\hat{m}}\right) + \lambda \cdot \mathcal{L}_{\text{Dice}}\left(\bm{m}, \bm{\hat{m}}\right)
\end{equation}
\noindent where $\mathcal{L}_{\text{BCE}}$ denotes the binary cross-entropy loss for pixel-wise classification, $\mathcal{L}_{\text{DICE}}$ is the Dice loss for region-level segmentation quality, and $\lambda$ is a weighting coefficient that balances their contributions, set to 0.5 by default.

\begin{figure*}[t]
    \centering
    \includegraphics[width=\linewidth]{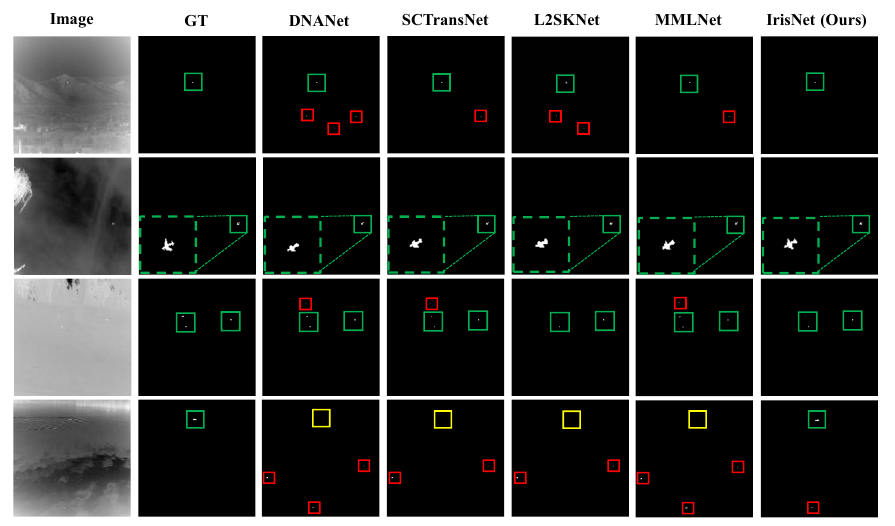} 
    \caption{Visual results of different IRSTD methods. The boxes in green, yellow, and red represent correct, missed, and false detections, respectively. The close-up views are shown in the corners with dashed lines.}         
    \label{fig:visual_results}            
\end{figure*}

\section{Experiments}
\label{sec:Experiments}
\subsection{Experimental Details}
\noindent \textbf{Datasets.}
We conduct extensive experiments on three public datasets: NUAA-SIRST~\cite{ACMNet}, NUDT-SIRST~\cite{DNANet}, and IRSTD-1K~\cite{ISNet}, containing 427, 1327, and 1000 images, respectively. These images cover a wide range of complex infrared scenarios. For NUAA-SIRST and NUDT-SIRST, we follow the 5:5 training/testing splits provided in~\cite{DNANet}, and for IRSTD-1K, we adopt the the 8:2 split proposed in~\cite{ISNet}. All dataset partitions are standard and widely used in the IRSTD community.

\noindent \textbf{Implementation Details.}
Our framework is implemented in PyTorch and trained on a single Nvidia GeForce RTX 4090 GPU. We use the Adam optimizer with a CosineAnnealing learning rate schedule. The initial learning rate is set to 8e-4 for NUAA-SIRST and IRSTD-1K, and 5e-4 for NUDT-SIRST. We train the model for a total of 800 epochs. Following~\cite{Sctransnet}, all images undergo the standard normalization and random cropping to generate $256 \times 256$ patches, and keep the original size of images for testing. We only adopt random flipping and cropping for data augmentation. The dimensions for multi-scale features $\{f^{(i)}\}_{i=1}^{5}$ are [32, 64, 128, 256, 512]. $\sigma$ is set to $5$. We use $6$ heads in the image-to-decoder transformer, with each head of $64$ hidden dimensions. The number of transformer layers is $l=6$.  

\subsection{Comparison with Other Methods}
\subsubsection{Quantitative Results.}
Table~\ref{tab:sota_1} summarizes the comparison between IrisNet and state-of-the-art IRSTD methods across three widely used datasets. Overall, IrisNet consistently achieves higher IoU and lower false alarm rates, with particularly strong gains on NUDT-SIRST and IRSTD-1K.

On IRSTD-1K, IrisNet attains a 72.08\% IoU, surpassing TCI-Former (70.14\%) and CFD-Net (69.97\%). Its 9.641 false alarm rate is also markedly lower than methods such as PSTNN and MSLSTIPT (35.26–1524), demonstrating robust performance under the dataset’s challenging low-SNR conditions.

For NUDT-SIRST, IrisNet outperforms all competitors by a significant margin, achieving 95.74\% IoU (1.65\% higher than SCTransNet at 94.09\%), a 99.37\% detection probability, and an optimal 0.367 of F$_{a}$. This highlights the effectiveness of IrisNet's meta-decoder in adapting to cross-scene variations, particularly in distinguishing between targets located in the sky or ground.

On NUAA-SIRST, IrisNet achieves 81.77\% IoU, outperforming several CNN-based methods such as CFD-Net (80.56\%) and GCI-Net (78.81\%). Notably, while GCI-Net and CFD-Net use larger input sizes (\textit{e.g.}, $512 \times 512$), IrisNet achieves competitive performance with a $256 \times 256$ input, demonstrating a better trade-off between accuracy and computational efficiency.

\subsubsection{Qualitative Results.}
Fig.~\ref{fig:visual_results} shows the detection results compared with several competitors in various complex scenarios. IrisNet stands out for its ability to achieve high detection accuracy and low false alarm rates. It outperforms many existing methods across all three datasets, especially in handling low-SNR conditions, providing a strong and efficient solution for infrared small-target detection.

\begin{table}[t]
  \centering
  \footnotesize 
    \setlength{\tabcolsep}{2mm}{
    \begin{tabular}{cccccccc}
    \toprule
    \multirow{2}{*}{Baseline} & \multirow{2}{*}{$\bm{x}_\text{hf}$} & \multicolumn{2}{c}{MKAB} & Meta & \multirow{2}{*}{IoU $\uparrow$} & \multirow{2}{*}{\textbf{$P_d \uparrow$}} & \multirow{2}{*}{\textbf{$F_a \downarrow$}} \\
    \cmidrule{3-4}
    & & MK & CA & Decoder & & &  \\
    \midrule
    \checkmark &  &  &  &  & 66.17 & 90.92 & 16.68  \\
    \checkmark &  &  &  & \checkmark & 68.38 & 91.25 & 16.51 \\
    \midrule
    \checkmark & \checkmark &  &  & \checkmark & 69.40 & 90.57 & 15.78 \\
    \checkmark & \checkmark & \checkmark & & \checkmark & 69.93 & 92.93 & 12.26 \\
    \checkmark & \checkmark & \checkmark & \checkmark & \checkmark & 72.08 & 92.59 & 9.641 \\
    \bottomrule
  \end{tabular}}
  \caption{Ablation study on key components of IrisNet. ``MK'' and ``CA'' denote the multi-kernel and channel attention designs in the multi-kernel aggregation block (MKAB).}
  \label{tab:ablation_study}
\end{table}

\begin{table}[t]
  \centering
  \footnotesize 
    \setlength{\tabcolsep}{3.5mm}{
    \begin{tabular}{lccc}
    \toprule
    \textsc{Methods} & IoU $\uparrow$ & $P_d \uparrow$ & $F_a \downarrow$ \\
    \midrule
    \textit{w/} Basic Decoder & 71.61 & 90.57 & 11.41 \\
    \textit{w/} Multi-scale Decoder & 72.28 & 90.57 & 8.919 \\
    \textit{w/} Spatial-attention Decoder & 72.08 & 92.59 & 9.641 \\
    \bottomrule
  \end{tabular}}
  \caption{Comparisons of different meta-decoder designs on the IRSTD-1K dataset.}
\label{tab:ablation_decoder}
\vspace{-0.2in}
\end{table}

\subsection{Ablation Study}

We provide in-depth studies to dissect the efficacy of different designs in our framework. For efficiency, we conduct ablation studies on IRSTD-1K dataset.

\noindent \textbf{Efficacy of Meta Decoder.}
Meta decoder allows the model to dynamically adapt its decoding strategy across different scenarios and target characteristics. However, a straightforward question is whether the meta decoder is truly necessary? To answer it, we first conduct an ablation study by directly learning the static decoder pattern or by using the image-to-decoder transformer to build the meta-learned decoding paradigm. 
Results are presented in the first two rows of Tab.~\ref{tab:ablation_study}. Introducing the meta-decoder leads to an IoU improvement of at least 2 points, with no increase in the false alarm ($F_{a}$) rate, and approximately a 0.5\% gain in $P_{d}$. These improvements strongly support both the motivation behind our approach and the effectiveness of the proposed framework.
Particularly, IRSTD-1K poses significant challenges due to its diverse imaging platforms (\textit{e.g.}, aerial and ground views) and complex backgrounds (\textit{e.g.}, buildings and grasslands). The dynamic decoding pattern helps the model adapt to varying image status, allowing for more precise localization of infrared targets.

\noindent \textbf{Impact of Designs in the Encoder.}
As previously discussed, a more powerful encoder can capture richer image status information, which is beneficial for guiding the image-to-decoder transformer to produce a decoding paradigm that is better aligned with the input. Therefore, we further explore several designs for the image encoder, including the integration of high-frequency information $\bm{x}_\text{hf}$, the usage of multi-kernel convolution, and a channel-attention strategy in the MKAB module. Results are presented in the last three rows of Tab.~\ref{tab:ablation_study}.
First of all, as expected, incorporating high-frequency information helps to strengthen target-position and scene-edge information in the spatial domain, which in turn supports the construction of a more effective meta-learned decoding paradigm. This leads to both a reduction in false alarms and an improvement in IoU at the pixel level.
Secondly, the multi-scale receptive field can better capture contextual information in the image and understand the relationship between the foreground target and the background. As a result, adding multi-kernel convolution layers is able to promote our IrisNet to effectively reduce the false alarm rate in the target masks produced by the meta-decoder.
Last but not least, the channel-attention mechanism is designed to filter out redundant information produced by the multi-kernel convolutional layer and better utilize the capability of multiple receptive fields. As shown in the last row of Tab.~\ref{tab:ablation_study}, this leads to significant improvements in both IoU and $F_{a}$, clearly demonstrating the efficacy of our design.

\noindent \textbf{Analysis of Meta-decoder Types.}
Moreover, we investigate three different types of the meta decoder, as illustrated in Fig.~\ref{fig:metadecoder}. 
Based on the results in Tab.~\ref{tab:ablation_decoder}, we can draw several key insights.
\textbf{(1)}
Thanks to our proposed decoder representation, IrisNet is compatible with three different meta-decoder structures, ranging from simple to more advanced designs. Even with a design where each decoder layer adopts a 2-branch, 4-layer-deep structure (\textit{i.e.}, spatial-attention decoder), our method can still establish an effective mapping from the image status to decoder configurations.
\textbf{(2)}
All three meta decoders performed well on the IRSTD-1K dataset. 
Compared with other approaches in Tab.~\ref{tab:sota_1}, the basic decoder already achieves the highest IoU score (71.61 \textit{vs.} 70.14), while its $F_{a}$ remained within an acceptable suboptimal range (11.41 \textit{vs.} 10.74).
\textbf{(3)} Generally, the effectiveness of the three meta-decoders follows: basic $<$ multi-scale $<$ spatial-attention. However, on the IRSTD-1K dataset, the multi-scale decoder shows a slight advantage over the spatial-attention. This also highlights the flexibility of our decoder representation, which can adaptively learn different decoding patterns based on the requirement.

\section{Conclusion}
\label{sec:Conclusion}

We present IrisNet, a meta-learned framework for infrared small target detection that addresses pattern drift across diverse scenarios. Using an image-to-decoder transformer, IrisNet dynamically maps infrared features to decoder parameters, improving robustness and adaptability. Its structured decoder captures inter-layer dependencies for stable, diverse decoding, while high-frequency components enhance target and edge perception. Experiments on multiple datasets demonstrate that IrisNet achieves state-of-the-art performance in complex infrared detection tasks.

{
    \small
    \bibliographystyle{ieeenat_fullname}
    \bibliography{main}

\begin{thebibliography}{44}
\providecommand{\natexlab}[1]{#1}
\providecommand{\url}[1]{\texttt{#1}}
\expandafter\ifx\csname urlstyle\endcsname\relax
  \providecommand{\doi}[1]{doi: #1}\else
  \providecommand{\doi}{doi: \begingroup \urlstyle{rm}\Url}\fi

\bibitem[Babu et~al.(2024)Babu, Liu, Zhou, Maire, Shakhnarovich, and Hanocka]{HyperFields}
Sudarshan Babu, Richard Liu, Avery Zhou, Michael Maire, Greg Shakhnarovich, and Rana Hanocka.
\newblock Hyperfields: Towards zero-shot generation of nerfs from text, 2024.

\bibitem[Bai and Zhou(2010)]{Top-hat}
Xiangzhi Bai and Fugen Zhou.
\newblock Analysis of new top-hat transformation and the application for infrared dim small target detection.
\newblock \emph{Pattern Recognition}, 43:\penalty0 2145--2156, 2010.

\bibitem[Chen et~al.(2014)Chen, Li, Wei, Xia, and Tang]{ALCM}
C.~L.~Philip Chen, Hong Li, Yantao Wei, Tian Xia, and Yuan~Yan Tang.
\newblock A local contrast method for small infrared target detection.
\newblock \emph{IEEE Transactions on Geoscience and Remote Sensing}, 52\penalty0 (1):\penalty0 574--581, 2014.

\bibitem[Chen et~al.(2022)Chen, Wang, and Tan]{TransSurvey}
Gao Chen, Weihua Wang, and Sirui Tan.
\newblock Irstformer: A hierarchical vision transformer for infrared small target detection.
\newblock \emph{Remote Sensing}, 14\penalty0 (14):\penalty0 3258, 2022.

\bibitem[Chen et~al.(2024)Chen, Tan, Chu, Wu, Liu, and Yu]{TCIFormer}
Tianxiang Chen, Zhentao Tan, Qi Chu, Yue Wu, Bin Liu, and Nenghai Yu.
\newblock Tci-former: Thermal conduction-inspired transformer for infrared small target detection.
\newblock In \emph{Proceedings of the AAAI Conference on Artificial Intelligence}, pages 1201--1209, 2024.

\bibitem[Chen and Wang(2022)]{ECCV22Hyper}
Yinbo Chen and Xiaolong Wang.
\newblock Transformers as meta-learners for implicit neural representations.
\newblock In \emph{European Conference on Computer Vision}, pages 170--187. Springer, 2022.

\bibitem[Dai et~al.(2021{\natexlab{a}})Dai, Wu, Zhou, and Barnard]{ACMNet}
Yimian Dai, Yiquan Wu, Fei Zhou, and Kobus Barnard.
\newblock Asymmetric contextual modulation for infrared small target detection.
\newblock In \emph{Proceedings of the IEEE/CVF winter conference on applications of computer vision}, pages 950--959, 2021{\natexlab{a}}.

\bibitem[Dai et~al.(2021{\natexlab{b}})Dai, Wu, Zhou, and Barnard]{ALCNet}
Yimian Dai, Yiquan Wu, Fei Zhou, and Kobus Barnard.
\newblock Attentional local contrast networks for infrared small target detection.
\newblock \emph{IEEE transactions on geoscience and remote sensing}, 59\penalty0 (11):\penalty0 9813--9824, 2021{\natexlab{b}}.

\bibitem[Deng et~al.(2021)Deng, Zhang, Xu, and Zhu]{TOPHAT2}
Lizhen Deng, Jieke Zhang, Guoxia Xu, and Hu Zhu.
\newblock Infrared small target detection via adaptive m-estimator ring top-hat transformation.
\newblock \emph{Pattern Recognition}, 112:\penalty0 107729, 2021.

\bibitem[Dosovitskiy et~al.(2020)Dosovitskiy, Beyer, Kolesnikov, Weissenborn, Zhai, Unterthiner, Dehghani, Minderer, Heigold, Gelly, et~al.]{ViT}
Alexey Dosovitskiy, Lucas Beyer, Alexander Kolesnikov, Dirk Weissenborn, Xiaohua Zhai, Thomas Unterthiner, Mostafa Dehghani, Matthias Minderer, Georg Heigold, Sylvain Gelly, et~al.
\newblock An image is worth 16x16 words: Transformers for image recognition at scale.
\newblock \emph{arXiv preprint arXiv:2010.11929}, 2020.

\bibitem[Fan et~al.(2024)Fan, Wang, Hu, Li, Dong, Zheng, Lin, Huang, and Ding]{DCFRNet}
Linyu Fan, Yingying Wang, Guoliang Hu, Feifei Li, Yuhang Dong, Hui Zheng, Changqing Lin, Yue Huang, and Xinghao Ding.
\newblock Diffusion-based continuous feature representation for infrared small-dim target detection.
\newblock \emph{IEEE Transactions on Geoscience and Remote Sensing}, 62:\penalty0 1--17, 2024.

\bibitem[Gao et~al.(2013)Gao, Meng, Yang, Wang, Zhou, and Hauptmann]{IPI}
Chenqiang Gao, Deyu Meng, Yi Yang, Yongtao Wang, Xiaofang Zhou, and Alexander~G. Hauptmann.
\newblock Infrared patch-image model for small target detection in a single image.
\newblock \emph{IEEE Transactions on Image Processing}, 22\penalty0 (12):\penalty0 4996--5009, 2013.

\bibitem[Ha et~al.(2016)Ha, Dai, and Le]{2016Hypernetworks}
David Ha, Andrew Dai, and Quoc~V Le.
\newblock Hypernetworks.
\newblock \emph{arXiv preprint arXiv:1609.09106}, 2016.

\bibitem[Hou et~al.(2022)Hou, Zhang, Tan, Xi, Zheng, and Li]{ISTDUNet}
Qingyu Hou, Liuwei Zhang, Fanjiao Tan, Yuyang Xi, Haoliang Zheng, and Na Li.
\newblock Istdu-net: Infrared small-target detection u-net.
\newblock \emph{IEEE Geoscience and Remote Sensing Letters}, 19:\penalty0 1--5, 2022.

\bibitem[Ioffe and Szegedy(2015)]{BN}
Sergey Ioffe and Christian Szegedy.
\newblock Batch normalization: Accelerating deep network training by reducing internal covariate shift, 2015.

\bibitem[Kang et~al.(2024)Kang, Ting, Ting, and Phan]{ASFYOLO}
Ming Kang, Chee-Ming Ting, Fung~Fung Ting, and Raphaël C.-W. Phan.
\newblock Asf-yolo: A novel yolo model with attentional scale sequence fusion for cell instance segmentation.
\newblock \emph{Image and Vision Computing}, 147:\penalty0 105057, 2024.

\bibitem[Kou et~al.(2023)Kou, Wang, Peng, Zhao, Chen, Han, Huang, Yu, and Fu]{NetSurvey}
Renke Kou, Chunping Wang, Zhenming Peng, Zhihe Zhao, Yaohong Chen, Jinhui Han, Fuyu Huang, Ying Yu, and Qiang Fu.
\newblock Infrared small target segmentation networks: A survey.
\newblock \emph{Pattern recognition}, 143:\penalty0 109788, 2023.

\bibitem[Li et~al.(2022{\natexlab{a}})Li, Xiao, Wang, Wang, Lin, Li, An, and Guo]{DNANet}
Boyang Li, Chao Xiao, Longguang Wang, Yingqian Wang, Zaiping Lin, Miao Li, Wei An, and Yulan Guo.
\newblock Dense nested attention network for infrared small target detection.
\newblock \emph{IEEE Transactions on Image Processing}, 32:\penalty0 1745--1758, 2022{\natexlab{a}}.

\bibitem[Li et~al.(2022{\natexlab{b}})Li, Li, Xiong, and Hoi]{BLIP}
Junnan Li, Dongxu Li, Caiming Xiong, and Steven Hoi.
\newblock Blip: Bootstrapping language-image pre-training for unified vision-language understanding and generation, 2022{\natexlab{b}}.

\bibitem[Li et~al.(2025)Li, Zhang, Lu, and Wang]{MMLNet}
Qiang Li, Wei Zhang, Wanxuan Lu, and Qi Wang.
\newblock Multibranch mutual-guiding learning for infrared small target detection.
\newblock \emph{IEEE Transactions on Geoscience and Remote Sensing}, 63:\penalty0 1--10, 2025.

\bibitem[Nirkin et~al.(2021)Nirkin, Wolf, and Hassner]{Hyperseg}
Yuval Nirkin, Lior Wolf, and Tal Hassner.
\newblock Hyperseg: Patch-wise hypernetwork for real-time semantic segmentation.
\newblock In \emph{Proceedings of the IEEE/CVF conference on computer vision and pattern recognition}, pages 4061--4070, 2021.

\bibitem[Oh and Peng(2022)]{Cvae-H}
Geunseob Oh and Huei Peng.
\newblock Cvae-h: Conditionalizing variational autoencoders via hypernetworks and trajectory forecasting for autonomous driving.
\newblock \emph{arXiv preprint arXiv:2201.09874}, 2022.

\bibitem[Park et~al.(2021)Park, Sinha, Hedman, Barron, Bouaziz, Goldman, Martin-Brualla, and Seitz]{HyperNeRF}
Keunhong Park, Utkarsh Sinha, Peter Hedman, Jonathan~T. Barron, Sofien Bouaziz, Dan~B Goldman, Ricardo Martin-Brualla, and Steven~M. Seitz.
\newblock Hypernerf: A higher-dimensional representation for topologically varying neural radiance fields, 2021.

\bibitem[Ronneberger et~al.(2015)Ronneberger, Fischer, and Brox]{UNet}
Olaf Ronneberger, Philipp Fischer, and Thomas Brox.
\newblock U-net: Convolutional networks for biomedical image segmentation, 2015.

\bibitem[Sun et~al.(2023)Sun, Bai, Yang, and Bai]{RDIAN}
Heng Sun, Junxiang Bai, Fan Yang, and Xiangzhi Bai.
\newblock Receptive-field and direction induced attention network for infrared dim small target detection with a large-scale dataset irdst.
\newblock \emph{IEEE Transactions on Geoscience and Remote Sensing}, 61:\penalty0 1--13, 2023.

\bibitem[Sun et~al.(2021)Sun, Yang, and An]{MSLSTIPT}
Yang Sun, Jungang Yang, and Wei An.
\newblock Infrared dim and small target detection via multiple subspace learning and spatial-temporal patch-tensor model.
\newblock \emph{IEEE Transactions on Geoscience and Remote Sensing}, 59\penalty0 (5):\penalty0 3737--3752, 2021.

\bibitem[Teutsch and Krüger(2010)]{sea_assist}
Michael Teutsch and Wolfgang Krüger.
\newblock Classification of small boats in infrared images for maritime surveillance.
\newblock In \emph{2010 International WaterSide Security Conference}, pages 1--7, 2010.

\bibitem[Wang et~al.(2022)Wang, Du, Liu, and Cao]{IAANet}
Kewei Wang, Shuaiyuan Du, Chengxin Liu, and Zhiguo Cao.
\newblock Interior attention-aware network for infrared small target detection.
\newblock \emph{IEEE Transactions on Geoscience and Remote Sensing}, 60:\penalty0 1--13, 2022.

\bibitem[Wu et~al.(2025)Wu, Liu, Zhang, Zhang, Luo, and Peng]{L2SKNet}
Fengyi Wu, Anran Liu, Tianfang Zhang, Luping Zhang, Junhai Luo, and Zhenming Peng.
\newblock Saliency at the helm: Steering infrared small target detection with learnable kernels.
\newblock \emph{IEEE Transactions on Geoscience and Remote Sensing}, 63:\penalty0 1--14, 2025.

\bibitem[Wu et~al.(2023)Wu, Li, Luo, Wang, Xiao, Liu, Yang, An, and Guo]{MTUNet}
Tianhao Wu, Boyang Li, Yihang Luo, Yingqian Wang, Chao Xiao, Ting Liu, Jungang Yang, Wei An, and Yulan Guo.
\newblock Mtu-net: Multilevel transunet for space-based infrared tiny ship detection.
\newblock \emph{IEEE Transactions on Geoscience and Remote Sensing}, 61:\penalty0 1–15, 2023.

\bibitem[Wu et~al.(2022)Wu, Hong, and Chanussot]{UIUNet}
Xin Wu, Danfeng Hong, and Jocelyn Chanussot.
\newblock Uiu-net: U-net in u-net for infrared small object detection.
\newblock \emph{IEEE Transactions on Image Processing}, 32:\penalty0 364--376, 2022.

\bibitem[Xiu et~al.(2023)Xiu, Ma, Pun, and Liu]{MDAFNet}
Xiaochen Xiu, Xianping Ma, Man-On Pun, and Ming Liu.
\newblock Mdafnet: Monocular depth-assisted fusion networks for semantic segmentation of complex urban remote sensing data.
\newblock In \emph{IGARSS 2023 - 2023 IEEE International Geoscience and Remote Sensing Symposium}, pages 6847--6850, 2023.

\bibitem[Ying et~al.(2023)Ying, Wang, Wang, Sheng, Liu, Lin, and Zhou]{publicsecurity}
Xinyi Ying, Yingqian Wang, Longguang Wang, Weidong Sheng, Li Liu, Zaiping Lin, and Shilin Zhou.
\newblock Local motion and contrast priors driven deep network for infrared small target super-resolution, 2023.

\bibitem[Yuan et~al.(2024)Yuan, Qin, Yan, Akhtar, and Mian]{Sctransnet}
Shuai Yuan, Hanlin Qin, Xiang Yan, Naveed Akhtar, and Ajmal Mian.
\newblock Sctransnet: Spatial-channel cross transformer network for infrared small target detection.
\newblock \emph{IEEE Transactions on Geoscience and Remote Sensing}, 62:\penalty0 1--15, 2024.

\bibitem[Zhang and Peng(2019)]{PSTNN}
Landan Zhang and Zhenming Peng.
\newblock Infrared small target detection based on partial sum of the tensor nuclear norm.
\newblock \emph{Remote Sensing}, 11\penalty0 (4), 2019.

\bibitem[Zhang et~al.(2022{\natexlab{a}})Zhang, Bai, Zhang, Zhang, Wang, Guo, and Gao]{Rkformer}
Mingjin Zhang, Haichen Bai, Jing Zhang, Rui Zhang, Chaoyue Wang, Jie Guo, and Xinbo Gao.
\newblock Rkformer: Runge-kutta transformer with random-connection attention for infrared small target detection.
\newblock In \emph{Proceedings of the 30th ACM International Conference on Multimedia}, pages 1730--1738, 2022{\natexlab{a}}.

\bibitem[Zhang et~al.(2022{\natexlab{b}})Zhang, Yue, Zhang, Li, and Gao]{FC3-Net}
Mingjin Zhang, Ke Yue, Jing Zhang, Yunsong Li, and Xinbo Gao.
\newblock Exploring feature compensation and cross-level correlation for infrared small target detection.
\newblock In \emph{Proceedings of the 30th ACM International Conference on Multimedia}, page 1857–1865, New York, NY, USA, 2022{\natexlab{b}}. Association for Computing Machinery.

\bibitem[Zhang et~al.(2022{\natexlab{c}})Zhang, Zhang, Yang, Bai, Zhang, and Guo]{ISNet}
Mingjin Zhang, Rui Zhang, Yuxiang Yang, Haichen Bai, Jing Zhang, and Jie Guo.
\newblock Isnet: Shape matters for infrared small target detection.
\newblock In \emph{Proceedings of the IEEE/CVF conference on computer vision and pattern recognition}, pages 877--886, 2022{\natexlab{c}}.

\bibitem[Zhang et~al.(2023{\natexlab{a}})Zhang, Zhang, Zhang, Guo, Li, and Gao]{Dim2Clear}
Mingjin Zhang, Rui Zhang, Jing Zhang, Jie Guo, Yunsong Li, and Xinbo Gao.
\newblock Dim2clear network for infrared small target detection.
\newblock \emph{IEEE Transactions on Geoscience and Remote Sensing}, 61:\penalty0 1--14, 2023{\natexlab{a}}.

\bibitem[Zhang et~al.(2024{\natexlab{a}})Zhang, Yang, Guo, Li, Gao, and Zhang]{IRPruneDet}
Mingjin Zhang, Handi Yang, Jie Guo, Yunsong Li, Xinbo Gao, and Jing Zhang.
\newblock Irprunedet: Efficient infrared small target detection via wavelet structure-regularized soft channel pruning.
\newblock In \emph{AAAI}, pages 7224--7232, 2024{\natexlab{a}}.

\bibitem[Zhang et~al.(2024{\natexlab{b}})Zhang, Yue, Li, Guo, Li, and Gao]{GCI-Net}
Mingjin Zhang, Ke Yue, Boyang Li, Jie Guo, Yunsong Li, and Xinbo Gao.
\newblock Single-frame infrared small target detection via gaussian curvature inspired network.
\newblock \emph{IEEE Transactions on Geoscience and Remote Sensing}, 62:\penalty0 1--13, 2024{\natexlab{b}}.

\bibitem[Zhang et~al.(2025)Zhang, Yue, Guo, Zhang, Zhang, and Gao]{CFD-Net}
Mingjin Zhang, Ke Yue, Jie Guo, Qiming Zhang, Jing Zhang, and Xinbo Gao.
\newblock Computational fluid dynamic network for infrared small target detection.
\newblock \emph{IEEE Transactions on Neural Networks and Learning Systems}, pages 1--13, 2025.

\bibitem[Zhang et~al.(2023{\natexlab{b}})Zhang, Li, Cao, Pu, and Peng]{AGPCNet}
Tianfang Zhang, Lei Li, Siying Cao, Tian Pu, and Zhenming Peng.
\newblock Attention-guided pyramid context networks for detecting infrared small target under complex background.
\newblock \emph{IEEE Transactions on Aerospace and Electronic Systems}, 59\penalty0 (4):\penalty0 4250--4261, 2023{\natexlab{b}}.

\bibitem[Zhao et~al.(2022)Zhao, Li, Li, Hu, Ma, and Tao]{IRSTDsurvey}
Mingjing Zhao, Wei Li, Lu Li, Jin Hu, Pengge Ma, and Ran Tao.
\newblock Single-frame infrared small-target detection: A survey.
\newblock \emph{IEEE Geoscience and Remote Sensing Magazine}, 10\penalty0 (2):\penalty0 87--119, 2022.

\end{thebibliography}
}

\end{document}